\documentclass[letterpaper, 10 pt, conference]{ieeeconf}  % Comment this line out if you need a4paper

\IEEEoverridecommandlockouts                              % This command is only needed if 
\usepackage{hyperref}
\hypersetup{
    colorlinks=true,
    linkcolor=black,
    filecolor=magenta,      
    urlcolor=cyan,
}
\usepackage[table]{xcolor}
\definecolor{maroon}{cmyk}{0,0.87,0.68,0.32}
\definecolor{Cyan}{HTML}{00F9DE}
\usepackage{graphicx} % for pdf, bitmapped graphics files
\graphicspath{ {IMAGES/} }

\usepackage{subfigure}
\usepackage{amsmath} % assumes amsmath package installed
\usepackage[lined,boxed,vlined]{algorithm2e}
\title{\LARGE \bf
FPGA based hybrid architecture for parallelizing RRT
}

\author{Gurshaant Malik, Krishna Gupta, Raunak Dharani and K Madhava Krishna% <-this % stops a space
}

\begin{document}

\maketitle
\thispagestyle{empty}
\pagestyle{empty}

%%%%%%%%%%%%%%%%%%%%%%%%%%%%%%%%%%%%%%%%%%%%%%%%%%%%%%%%%%%%%%%%%%%%%%%%%%%%%%%%
\begin{abstract}

Field Programmable Gate Arrays(FPGA) exceed the computing power of software based implementations by breaking the paradigm of sequential execution and accomplishing more per clock cycle by enabling hardware level parallelization at an architectural level. Introducing parallel architectures for a computationally intensive algorithm like Rapidly Exploring Random Trees(RRT) will result in an exploration that is fast, dense and uniform. Through a cost function delineated in later sections, FPGA based combinatorial architecture delivers superlative speed-up but consumes very high power while hierarchical architecture delivers relatively lower speed-up with acceptable power consumption levels. To combine the qualities of both, a hybrid architecture, that encompasses both combinatorial and hierarchical architecture, is designed. To determine the number of RRT nodes to be allotted to the combinatorial and hierarchical blocks of the hybrid architecture, a cost function, comprised of fundamentally inversely related speed-up and power parameters, is formulated. This maximization of cost function, with its associated constraints, is then mathematically solved using a modified branch and bound, that leads to optimal allocation of RRT modules to both blocks. It is observed that this hybrid architecture delivers the highest performance-per-watt out of the three architectures for differential, quad-copter and fixed wing kinematics. 

\end{abstract}

%%%%%%%%%%%%%%%%%%%%%%%%%%%%%%%%%%%%%%%%%%%%%%%%%%%%%%%%%%%%%%%%%%%%%%%%%%%%%%%%
\section{INTRODUCTION}

During the last decade and a half, as computer power has increased, sampling-based path planning algorithms, such as rapidly exploring random trees (RRT), have been shown to work well in practice and possess theoretical guarantees such as probabilistic completeness. A significant amount of research effort has gone into improving the performance of RRTs. From an architectural standpoint, recent research efforts have been directed towards parallelizing RRT \cite{devaurs2011parallelizing} \cite{stradfordscalable} \cite{c_malik2015fpga} \cite{malik2015fpga}. Out of these, distributed RRT \cite{devaurs2011parallelizing} proffers the use of MPI for inter-module communication between multiple RRT modules to maintain data sanity, at the cost of inter-RRT scheduling. K-distributed \cite{stradfordscalable} reduces this scheduling by lowring the amount of inter-RRT communication, at the cost of a less uniform exploration. However, FPGA based combinatorial \cite{c_malik2015fpga} and hierarchical \cite{malik2015fpga} architectures, have already been shown to perform better than these implementations. FPGA enables delivery of tightly packed, energy efficient infrastructures adept in fast real time performance \cite{fischer2011rotation} \cite{svab2009fpga} \cite{oberg2012random} \cite{dubey2012field}. Unlike a software effort in parallelization \cite{devaurs2011parallelizing} \cite{stradfordscalable}, an FPGA allows gate level control of system architecture for parallelizing RRT. This allows the designer to tap the potential of hardware design, allowing control over minute details of arithmetic design, real time parallelization, pipe-lining of sequential processes. The hardware level flexibility afforded by an FPGA results in parallel RRT architectures that are not only fast, but also small and power efficient.

\begin{figure}[h]
    \includegraphics[width=8.6cm, height=10.2cm]{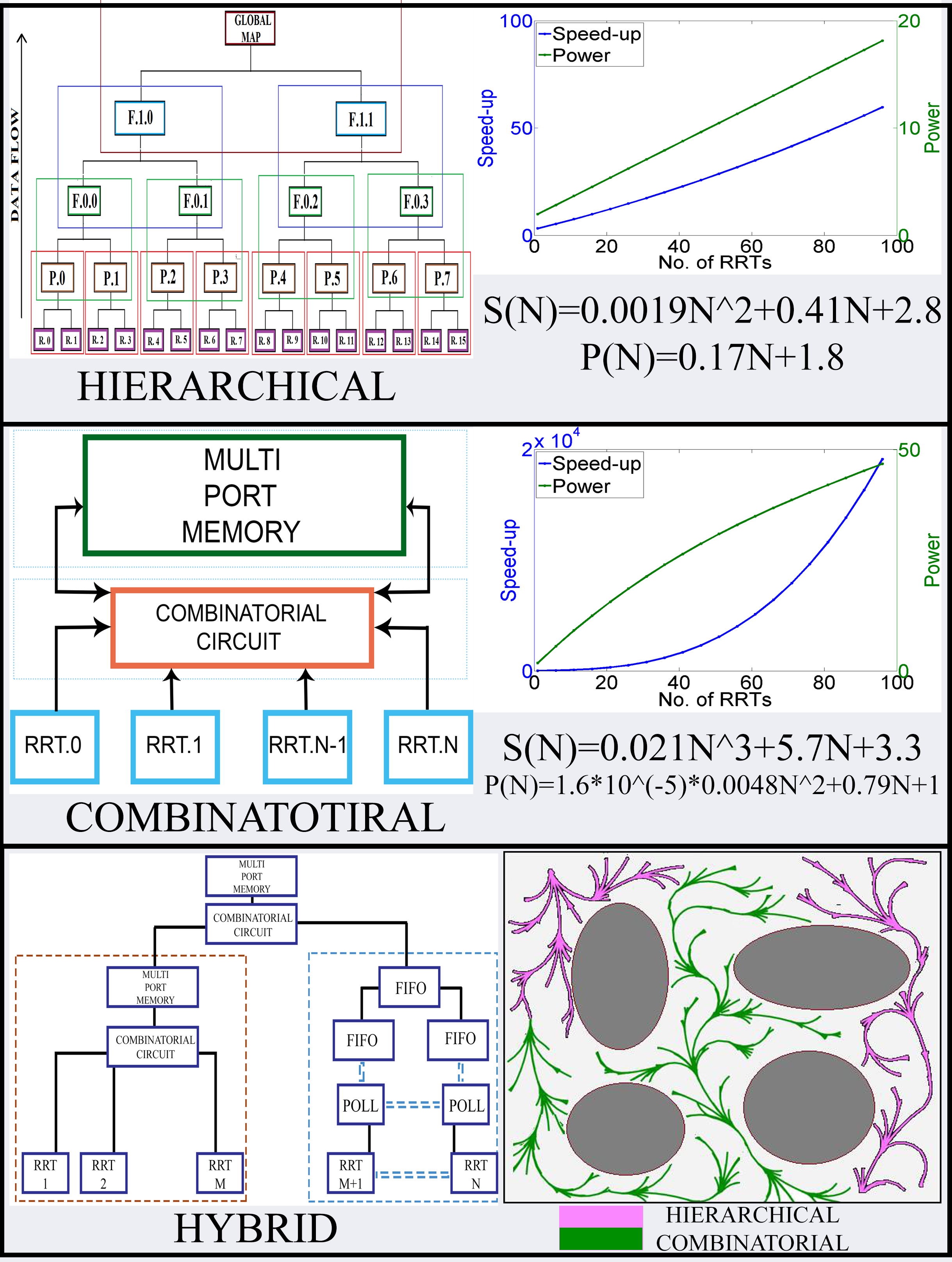}
    %\centering
    \caption{Overview : Hierarchical, Combinatorial and Hybrid architectures}
  %  \vspace{-3.5em}
    \label{Introduction_Overview_Architectures}
\end{figure}

FPGA based combinatorial and hierarchical architectures have already been shown to perform better than state of the art parallel RRT architectures like distributed \cite{devaurs2011parallelizing} and K-distributed \cite{stradfordscalable}. Fig. \ref{Introduction_Overview_Architectures} summarises the respective speed-up and power consumption, as a function of $N$ parallel RRTs for combinatorial and hierarchical architectures. As shown, the speed-up and power consumption of combinatorial is magnitudes larger than hierarchical. Theoretically, an architecture with maximum speed-up and minimum power consumption is desired. To converge towards this theoretical ideality, as shown in Fig. \ref{Introduction_Overview_Architectures}, a flexible and malleable hybrid architecture that consists of $M$ RRTs allotted to combinatorial and $N-M$ RRTs allotted to hierarchical is proposed. The determination of $M$ is mathematically calculated, with the calculations centered around maximization of a cost function, using a set of constraints explained in later sections. The subsequently designed hybrid architecture is then tested successfully for scalability across robotic kinematic complexity and geometric complexity by deploying it for a differentially driven system, a quad-copter and fixed wing aircraft in geometrically constrained environments.

\section{CHALLENGES IN PARALLELIZING RRT}

\begin{figure}[h]
    \centering
    \includegraphics[scale=0.2]{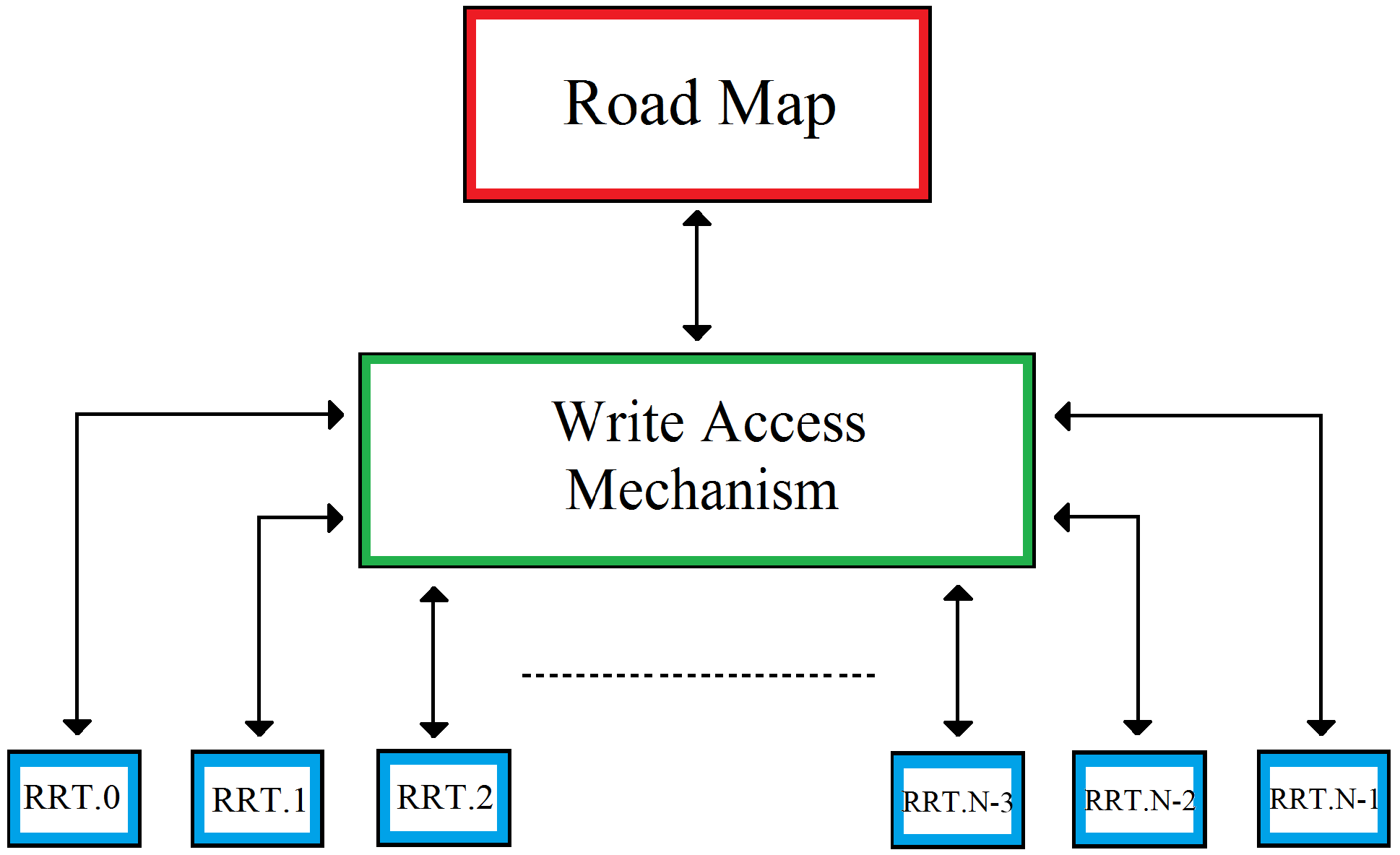}
    \caption{Schematic illustration of exploratory decomposition}
    %\vspace{-2em}
    \label{Exploratory_Decomposition}
\end{figure}

Since RRT involves randomized exploration of the map, ours and many proposed algorithms \cite{devaurs2011parallelizing} \cite{stradfordscalable} use the principle of exploratory decomposition \cite{kumar1994introduction} as their foundation. In other words, each RRT produces its own output and through different write mechanisms, the outputs are integrated to build the explored map. Fig. \ref{Exploratory_Decomposition} provides an overview of this design philosophy.

\begin{figure}[h]
    \centering
    \includegraphics[scale=0.2]{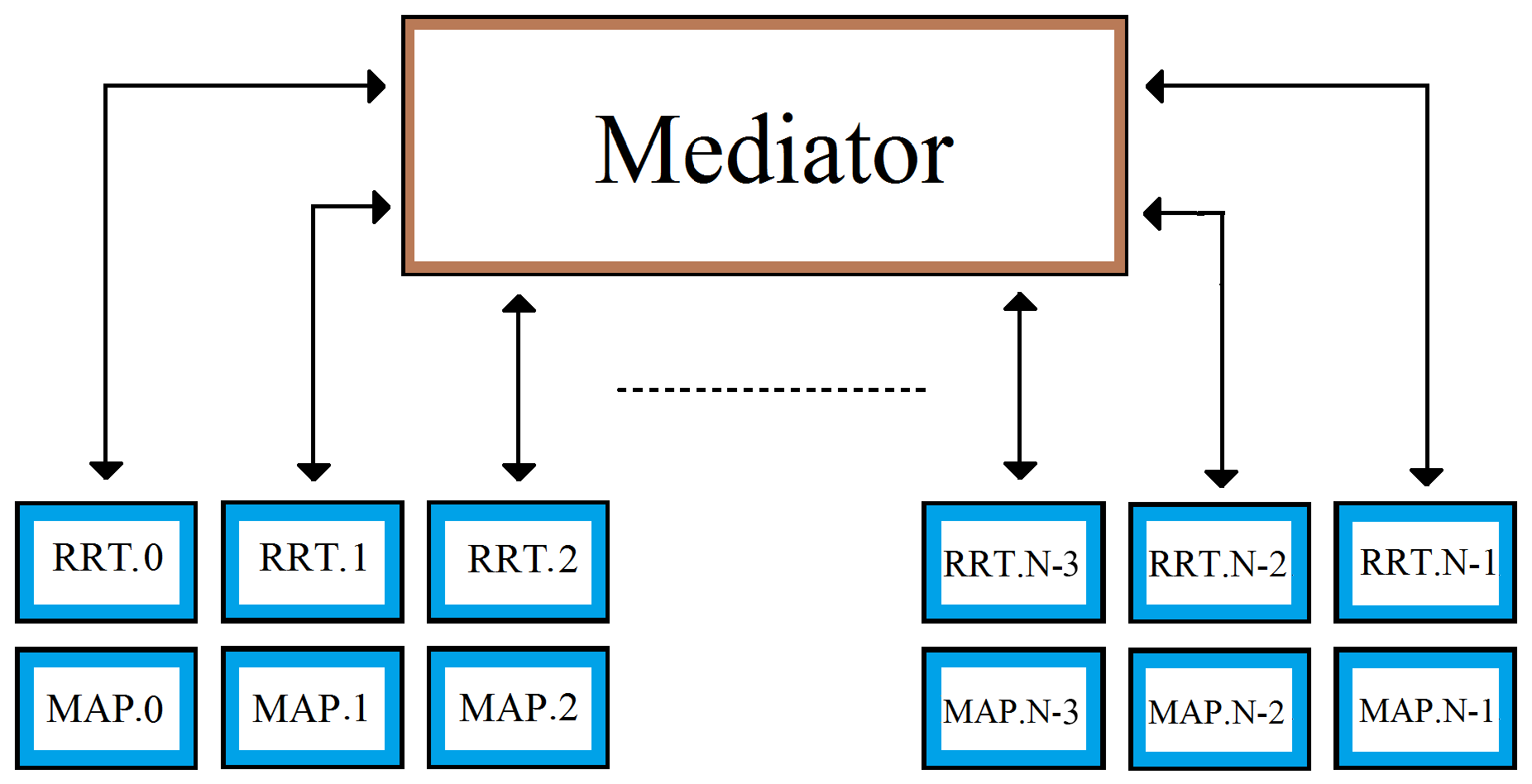}
    \caption{Schematic illustration of distributed philosophy}
    %\vspace{-2em}
    \label{Distributed_Philosophy}
\end{figure}

In such a parallel RRT design, an important issue is to decide the write access mechanism that integrates the data from multiple RRTs and then updates the global explored map. There are 2 general philosophies : 1.) Distributed and 2.) Shared. The distributed philosophy employs a scheme by which each RRT will have its own local explored map. As a result, changes made by it to its local explored map will have no effect on other RRT's local explored maps. Hence, as shown in Fig. \ref{Distributed_Philosophy}, we need a ’mediator’ system that updates each RRT's local explored map to changes made by other RRTs. This will incur significant inter-RRT communication time in case of large scale parallelization.

\begin{figure}[h]
    \centering
    \includegraphics[scale=0.17]{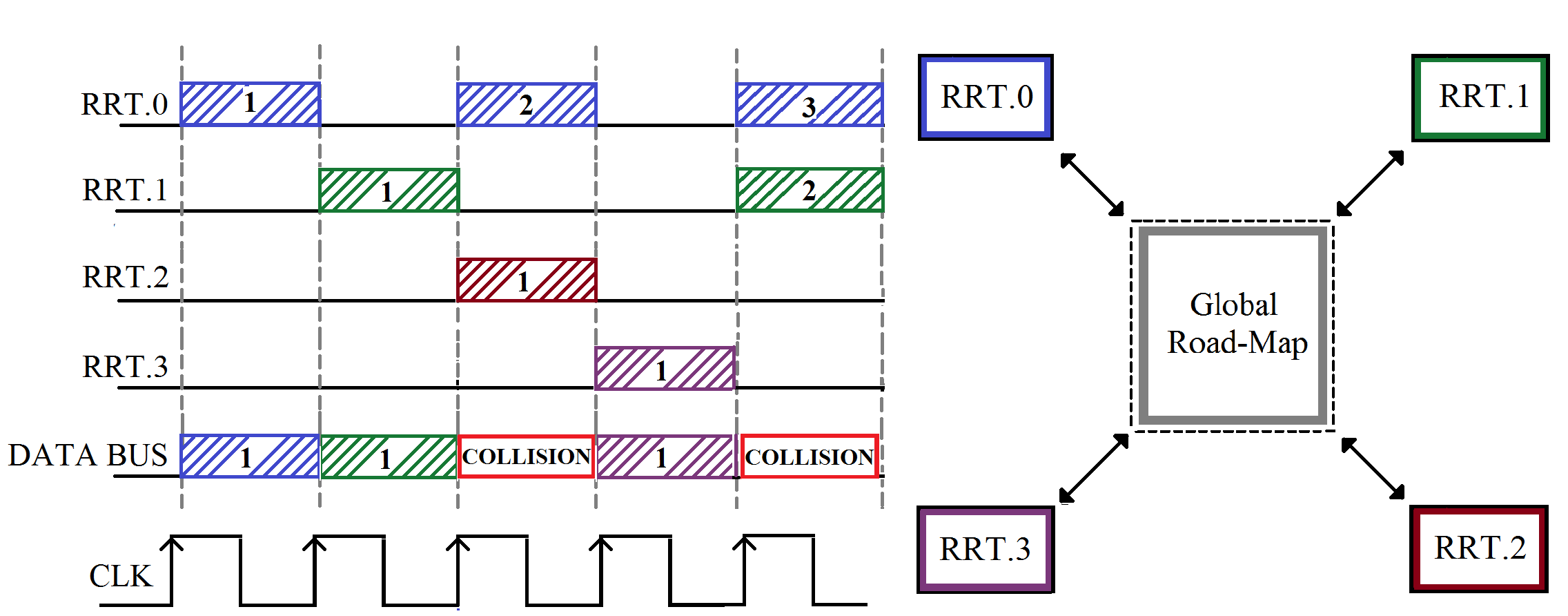}
    \caption{Schematic illustration of the Shared design philosophy}
    %\vspace{-.5em}
    \label{Shared_Philosophy}
\end{figure}

The shared design philosophy, shown in Fig. \ref{Shared_Philosophy}, allows all the RRTs to have access to the same global explored map. Hence, there is virtually no inter-RRT communication. But, since all RRTs will have access to the same global explored map, large scale parallelization, without scheduling, can geometrically increase traffic on global address space, leading to data collisions.

\begin{figure}[ht]
    \centering
    \includegraphics[width=8.6cm, height=3.5cm]{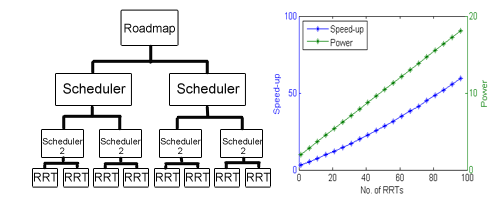}
    \caption{Hierarchical : Speed-up and Power plots. Best at 120\% zoom}
    %\vspace{-.5em}
    \label{Hierarchical}
\end{figure}

Conventional software implementation of parallel RRTs \cite{devaurs2011parallelizing} \cite{stradfordscalable} solve the problem of this contentious relationship between scheduling and data integrity by using MPI, STAPLE frameworks. However hardware implementations, owing to RTL level optimizations being impossible on conventional software, have been shown to perform significantly better than their software counterparts in the case of parallel RRTs \cite{c_malik2015fpga} \cite{malik2015fpga}. As shown in Fig. \ref{Hierarchical}, FPGA based hierarchical architecture proposes a binary tree, that combines shared and distributed memories, limits inter RRT scheduling to sibling RRTs only. Hence, it has a respectable speed and power cost function. 

\begin{figure}[ht]
    \centering
    \includegraphics[width=8.6cm, height=3.5cm]{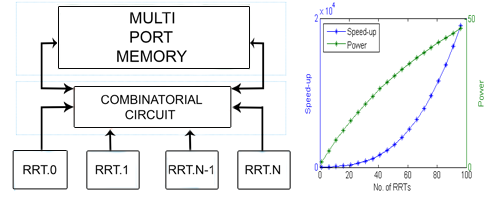}
    \caption{Combinatorial : Speed-up and Power plots. Best at 120\% zoom}
    %\vspace{-.5em}
    \label{Combinatorial}
\end{figure}

As shown in Fig. \ref{Combinatorial}, combinatorial architecture eliminates scheduling by introducing a multi-port shared memory and combinatorial multiplexing, possible only as an FPGA implementation, taking care of all the possible $2^N$ cases during a write window for a $N$ RRT system. Hence it has a very high speed-up but also a very high power cost function.

Ideally, a parallel RRT architecture should have a speed-up capability similar to combinatorial and power consumption levels similar to hierarchical. The next section delineates on this requirement with the proposed hybrid architecture.

\section{PROPOSED HYBRID ARCHITECTURE}

To enable intelligible understanding of the proposal, this section is further divided into 3 subsections. 1.) Hybrid Architecture's hypothesis, 2.) Hybrid Architecture's design and mathematical variables, 3.) Cost Function, to calculate the hybrid architecture's variables, strictly constrained by a set of intelligent, FPGA platform sensitive conditions.

\subsection{Hypothesis}

Owing to the probability reliant exploration of RRT, accurate prognosis of data arrival time is an ambiguous task. Hence $N$ RRTs working in parallel can result in $2^N$ possible cases during a write window to the global road-map. In order to theoretically rationalise the hypothesis behind the hybrid architecture, it is important to characterize the architecture, speed-up and power consumption levels of the hierarchical and combinatorial architectures, in chronological order.

\begin{figure}[h]
    \centering
    \includegraphics[width=8.6cm, height=3.5cm]{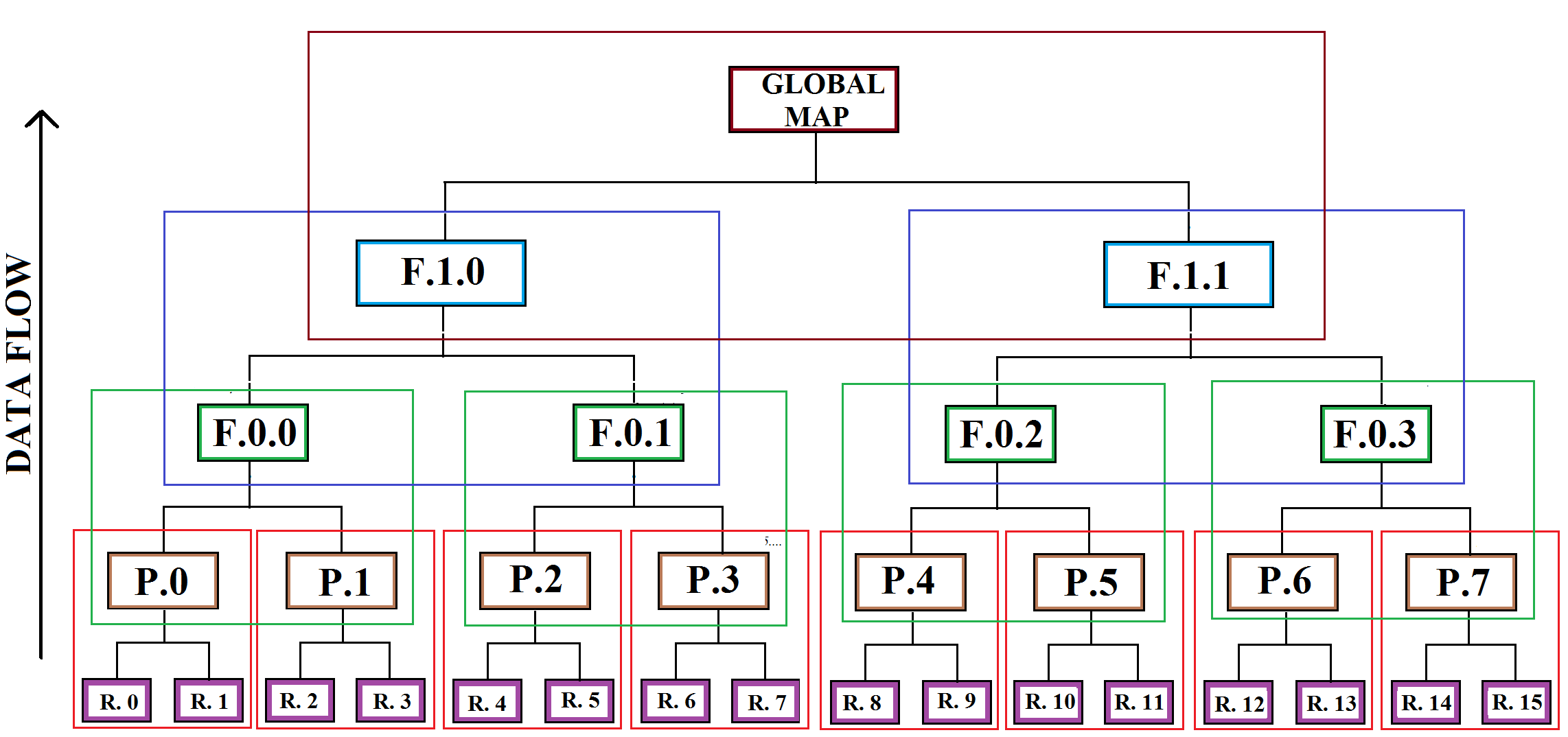}
    \caption{Hierarchical architecture}
    %\vspace{-.5em}
    \label{Hierarchical_integrated}
\end{figure}

In hierarchical, as shown in Fig. \ref{Hierarchical_integrated}, the data stems from the RRT modules and flows through higher levels of hierarchy to reach the global map. P stands for POLL, F stands for FIFO. At the deepest level, P0 chronologically polls RRT0, RRT1. P1 polls RRT2, RRT3 and so on. Going up, F00 polls P0, P1. F01 polls P2, P3 and so on. Going up a level, F10 polls F00, F01 and F11 polls F02, F03. Finally, at the highest level, the global road-map is updated by F10 and F11. At all levels, chronological polling for data by parent module preserves data integrity but the architecture is still weighed down by the scheduling that prevails amongst child modules of the parent modules. Eqs. \ref{Speed-Up_Hierarchical} and \ref{Power_Hierarchical} present the speed-up and power consumption levels respectively for the hierarchical architecture, extracted out of the data for speed-up and power available with the authors, for $N$ parallel RRT modules.

\begin{equation}
S(N)=0.0019N^2+0.41N+2.8
\label{Speed-Up_Hierarchical}
\end{equation}

\begin{equation}
P(N)=0.17N+1.8
\label{Power_Hierarchical}
\end{equation}

\begin{figure}[h]
    \includegraphics[width=8.6cm, height=5cm]{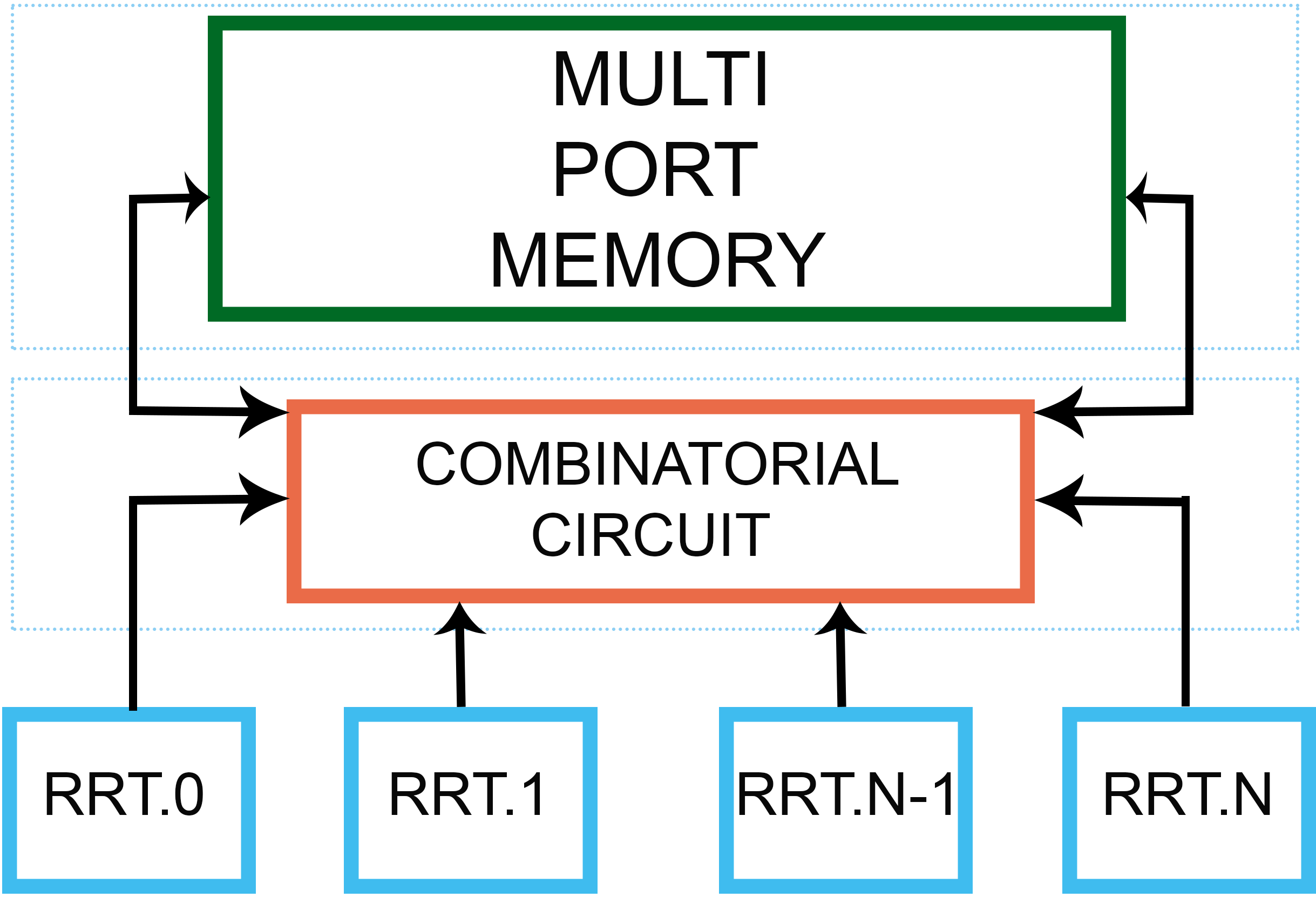}
    \caption{Combinatorial architecture}
    \label{Combinatorial_Integrated}
\end{figure}

In combinatorial, as shown in Fig. \ref{Combinatorial_Integrated}, the first part of the architecture is a multi-port random access memory with the ability to handle [$0$, $N$] variable, asynchronous write and/or read transactions during a write and/or read window. The second part of the architecture is a combinatorial circuit that ascertains the current case of the $2^N$ cases during the write window and feeds the appropriate write control signals to the memory. This allows each of the $N$ RRTs to have access to the write window with zero latency/scheduling since this write access mechanism combinatorially accounts for all the possible $2^N$ cases. Eqs. \ref{Speed-Up_Combinatorial} and \ref{Power_Combinatorial} present the speed-up and power consumption levels respectively for the combinatorial architecture, extracted out of the data for speed-up and power available with the authors, for $N$ parallel RRT modules.

\begin{equation}
S(N)=0.021N^3+5.7N+3.3
\label{Speed-Up_Combinatorial}
\end{equation}

\begin{equation}
P(N)=1.6*10^{-5}*N^3-0.0048N^2+0.79N+1
\label{Power_Combinatorial}
\end{equation}

\begin{figure}[h]
    \centering
    \includegraphics[width=8.6cm, height=4.6cm]{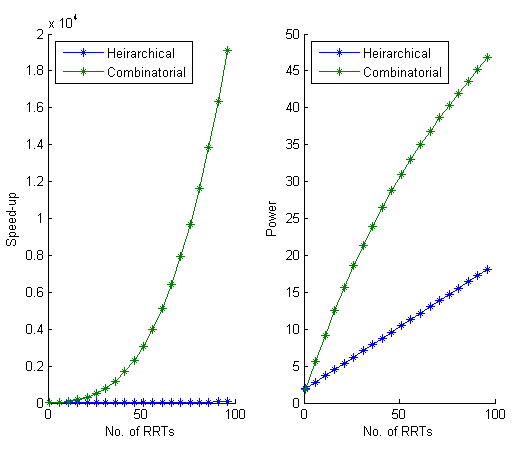}
    \caption{Speed-Up and Power : Combinatorial VS Hierarchical}
    %\vspace{-.5em}
    \label{Comparison_Speed-Up_Area}
\end{figure}

Comparing Eqs. \ref{Speed-Up_Hierarchical} and \ref{Speed-Up_Combinatorial}, also plotted in Fig. \ref{Comparison_Speed-Up_Area}, architecturally, combinatorial aggressively out-throttles hierarchical. But on the other hand, comparing Eqs. \ref{Power_Hierarchical} and \ref{Power_Combinatorial}, hierarchical is of a much more clement nature in power consumption. It should be noted that speed-up directly controls the accelerated capability of the system, that is, how fast a map is explored. Mobile robots typically are constrained by a small battery. Hence, quantitatively, a maximal bound that is very small in magnitude needs to be placed on power consumption levels. Theoretically, it can be concluded with confidence that an architecture that behaviorally is analogous to hierarchical in terms of power consumption and analogous to combinatorial in terms of speed-up is ideal. Hence, a hybrid architecture that combines these properties of hierarchical and combinatorial is hypothesized.

\subsection{Hybrid Architecture}

\begin{figure}[h]
    \includegraphics[width=8.6cm, height=4.9cm]{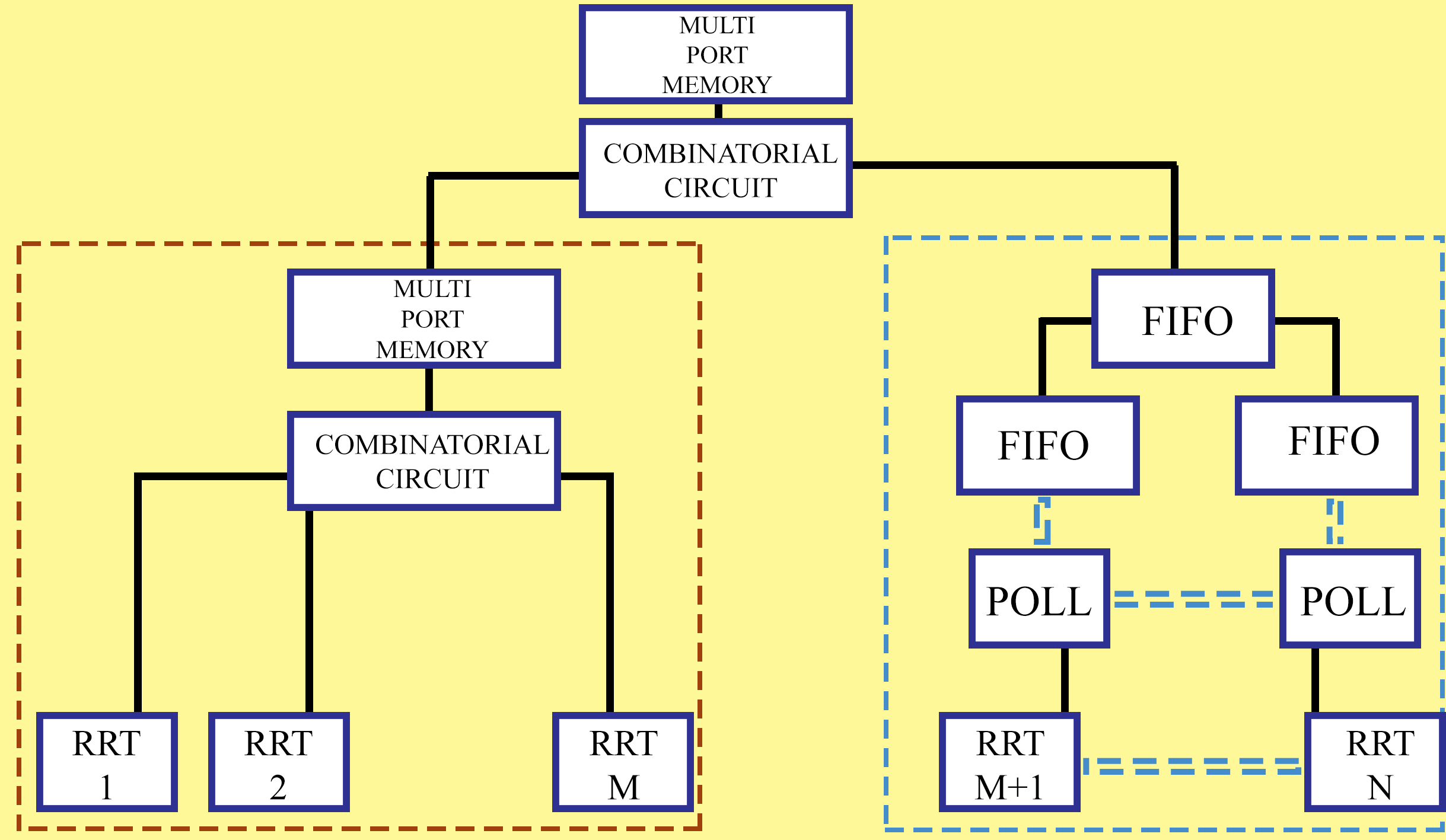}
    \caption{Schematic illustration of hybrid architecture}
    \label{Hybrid_Architecture}
\end{figure}

Fig. \ref{Hybrid_Architecture} provides a top level architectural view of the hybrid architecture for $N$ parallel RRT modules. Consequent delineation follows. The critical design philosophy instructs the division of these $N$ RRT modules into 2 parts : 1.) $M$ RRT modules are aligned to follow the combinatorial architecture and 2.) Remaining $N-M$ RRT modules are aligned to hierarchical architecture. Hence, by varying $M$, the  architecture can be made to cover the entire behavioral spectrum, with the extreme being hierarchical for $M=0$ and combinatorial for $M=N$. Eqs. \ref{Range_Hybrid_Speed-Up} and \ref{Range_Hybrid_Area} mathematically quantify this property.

\begin{equation}
S_{Hybrid}(M)\rightarrow[S_{Hier}(M=0),S_{Combi}(M=N)]
\label{Range_Hybrid_Speed-Up}
\end{equation}

\begin{equation}
P_{Hybrid}(M)\rightarrow[P_{Hier}(M=0),P_{Combi}(M=N)]
\label{Range_Hybrid_Area}
\end{equation}

\begin{equation}
Total_{Hier}(M)=N-M
\label{Number_Hierarchical}
\end{equation}

\begin{equation}
Total_{Combi}(M)=M+1
\label{Number_Combinatorial}
\end{equation}

As shown in Fig. \ref{Hybrid_Architecture}, $M$ RRT modules explore the map in parallel via combinatorial architecture and the remaining $N-M$ RRT modules explore the map in parallel via hierarchical architecture. These 2 exploration results are then combined together to form the  global explored road-map via combinatorial block(Hence, total $M+1$ blocks). Eqs. \ref{Number_Hierarchical} and \ref{Number_Combinatorial} describe the number of combinatorial and hierarchical blocks respectively. For a given value of $N$, the entire behavior of this architecture is administered by the variable $M$. The next subsection outlines the mathematical formulas and constraints critical to the deduction of the variable $M$. 

\subsection{Cost Function}

Owing to the mathematical complexity and volume involved, for unabridged understanding of the concept, this section is further divided into 4 subsections : 1.) The cost function, 2.) Set of constraints to generate a singular, optimized solution, 3.) The computation strategy to generate the solution and 4.) Arbitration of the tagging of $M$ RRT modules out of the total $N$ RRT modules. 

\subsubsection{Cost Function}

Eqs. \ref{Speed-Up_Hierarchical}, \ref{Power_Hierarchical}, \ref{Speed-Up_Combinatorial} and \ref{Power_Combinatorial}, clearly manifest the irreconcilable nature of maximality and minimality between speed-up and power consumption since both of them are proportional to the number of parallel RRT modules. That is, speed-up cannot be maximised in conjunction with minimised power consumption. Hence, instead of solitary maximization of speed-up and minimization of power consumption, we aim to maximise the cost function given in Eq. \ref{Cost_Function_Total}, with individual terms delineated in Eqs. \ref{Speed-Up_Total} and \ref{Area_Total}, for a particular value of $M$. While adjudicating about the formulation of the cost function, it was observed that the form $S/P$ was biased towards minimising power whereas $S+1/P$ was moderate in nature. As described in Eq. \ref{Number_Combinatorial}, it should be noted that for $M$ combinatorial RRT modules, there exist $M+1$ combinatorial blocks. 

\begin{equation}
%\scriptsize
J_{Hyb}(M)=S_{Total}(M)+\frac{1}{P_{Total}(M)}
\label{Cost_Function_Total}
\end{equation}

\begin{equation}
%\scriptsize
S_{Total}(M)=S_{Hier}(N-M)+S_{Combi}(M)
\label{Speed-Up_Total}
\end{equation}

\begin{equation}
%\scriptsize
P_{Total}(M)=P_{Hier}(N-M)+P_{Combi}(M+1)
\label{Area_Total}
\end{equation}

%\begin{equation}
%\scriptsize
%argmax(J_{Hyb}(N,M))=(M|max(J_{Hyb}(M)))
%\label{M}
%\end{equation}

\subsubsection{Set Of Constraints}
The cost function is maximised subject to the following constraints :

\begin{itemize}
    \item Naturally, $M$ must be a positive integer
    \begin{equation}
        M > 0, \in I
    \end{equation}
    
    \item $M$ must not exceed $N$
    \begin{equation}
        M \leq N
    \end{equation}
    
    \item Sensitive to robotic platform's battery endurance capability, the designer decides how much maximum power($\omega$) the architecture can consume. 
    \begin{equation}
        P_{Hyb}(M) \leq \omega
    \end{equation}
    
\end{itemize}

\subsubsection{Mathematical Solver}
To solve for $M$, we aim to maximise the cost function, as previously described in Eq. \ref{Cost_Function_Total}. For this, Branch and Bound \cite{lawler1966branch}, a systematic solver for optimized integer solutions, is used. Branch and bound has the ability to accept non-linear optimization problems as inputs, as is the case with the formulae. The generic algorithm of branch and bound follows.

\begin{figure}[h]
    \includegraphics[width=8.6cm, height=4cm]{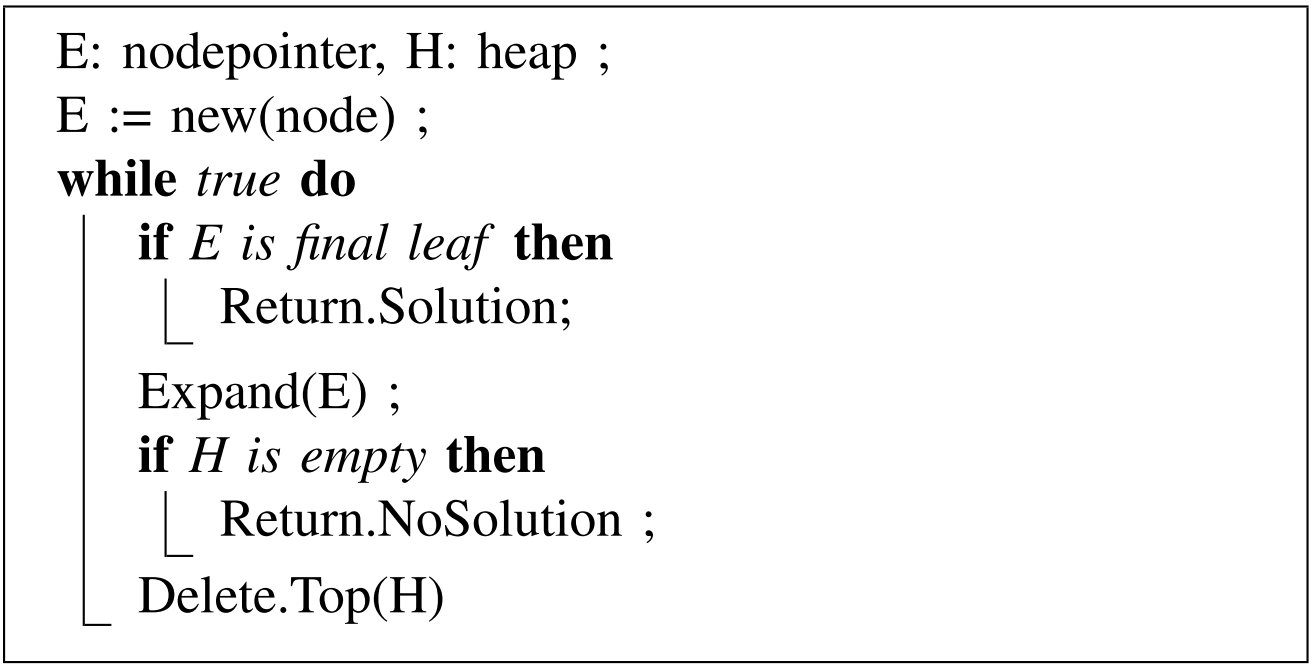}
\end{figure}
\vspace{2cm}

\subsubsection{Tagging of M RRT modules}
\begin{figure}[h]
    \includegraphics[width=8.4cm, height=5.8cm]{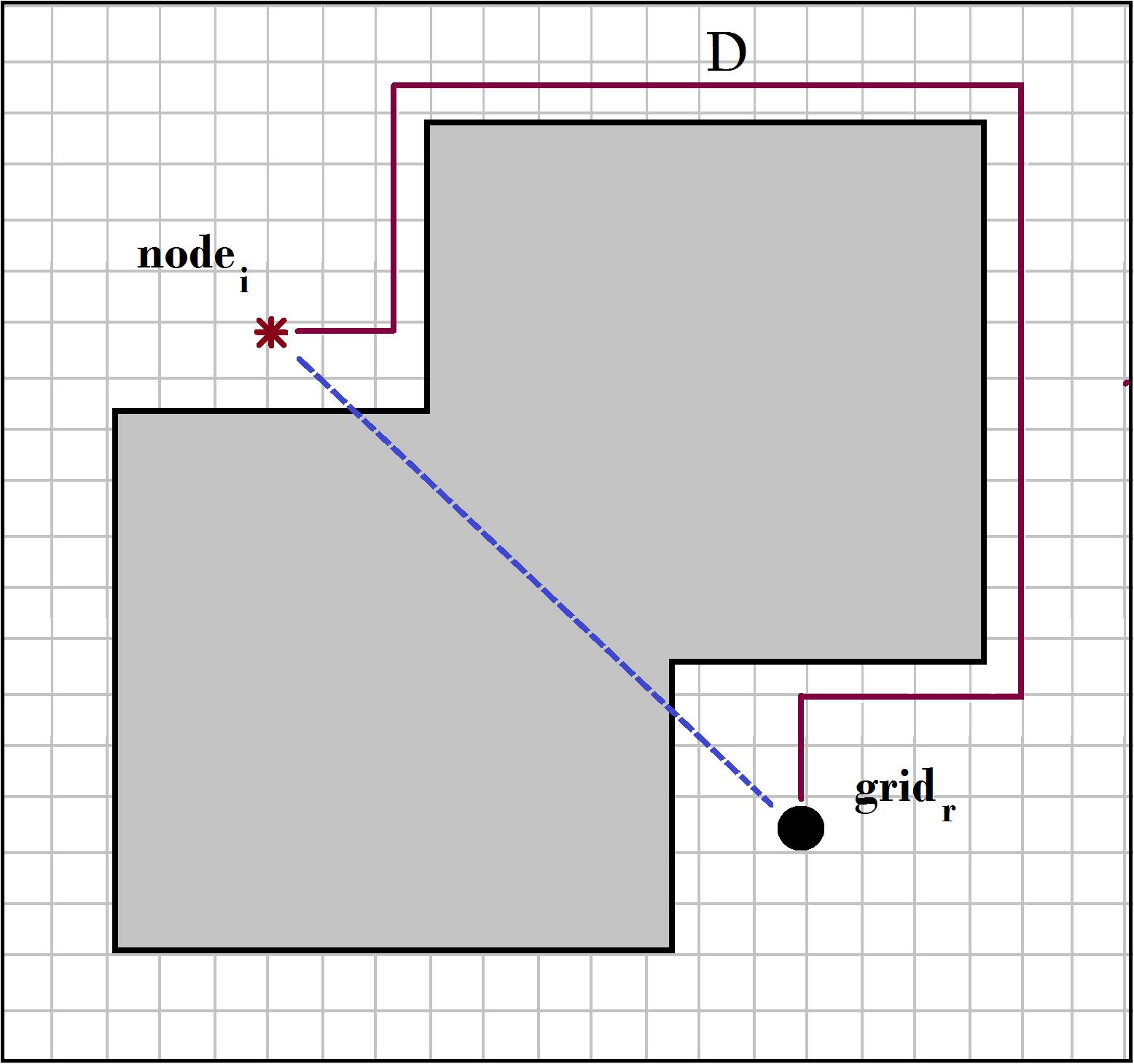}
    \caption{Calculation of distance for area estimation}
    \label{Final_Condition_Figure}
\end{figure}

Post the calculation of $M$, the next step assigns labels to each RRT as a combinatorial (M such RRTs) or hierarchical RRT ((N-M) such RRTs). This identification, as described in Eq. \ref{Final_Condition}, is driven by the user's decision about the approximate average map area($\alpha$) each of the combinatorial $M$ RRT modules should explore. As shown in Fig. \ref{Final_Condition_Figure}, the map is divided into high resolution grids and the area is calculated, as described in Eqs. \ref{Area} and \ref{Distance}, by summing the distance between that RRT module's starting node and each grid. It should be noted that the user has the flexibility of intelligently \cite{KMadhava} or randomly choosing the starting nodes. For the current experimental setup, a value of $\alpha$ was so chosen that the RRT nodes with the top $M$ areas were allotted to combinatorial architecture.

\begin{equation}
%\scriptsize
\sum_{i=1}^{M}\frac{A_{Comb}(i)}{M}\geq \alpha
\label{Final_Condition}
\end{equation}

\begin{equation}
%\scriptsize
A_{Comb}(i)=\frac{1}{\sum_{r=1}^{grids}d_{r}}*A_{Map}
\label{Area}
\end{equation}

\begin{equation}
%\scriptsize
d_{r}= BFS.distance(grid_{r}, node_{i})
\label{Distance}
\end{equation}

\section{FPGA Implementation}

The design platform, Zedboard, uses the Zynq-7000 SOC, with system parameters given in table below. For our design, the cartesian coordinates are represented as 32bit long, fixed-point, 2's compliment binary strings where the 24 MSB represent the integer part and the 8 LSB represent the fractional part. This representation provides an incremental resolution of 0.00390625 in decimal format. The geometrical angle is represented as a 16bit long, fixed point, 2s compliment binary string where the 3 MSB represent the integer part and the 13 LSB represent the fractional part, affording an incremental resolution of 0.00012207 radians.

\begin{table}[ht]
\centering
\label{tab:title}
\begin{tabular}{|l|l|l|l|l|}
\hline
\multicolumn{5}{|c|}
{\textbf{System Parameters}}                                        \\ \hline
\textbf{LUT}         & 17,600 & \textbf{Total BRAM(Mb)} & \multicolumn{2}{l|}{2.1}  \\ \hline
\textbf{Logic Cells}    & 2800 & \textbf{DSP48E1}        & \multicolumn{2}{l|}{80}   \\ \hline
\textbf{CLB Flip Flops} & 35,200 & \textbf{Area($inch^2$)}       & \multicolumn{2}{l|}{6.5*5.9} \\ \hline
\end{tabular}
\end{table}

\begin{figure}[h]
    \includegraphics[width=8.6cm, height=5.2cm]{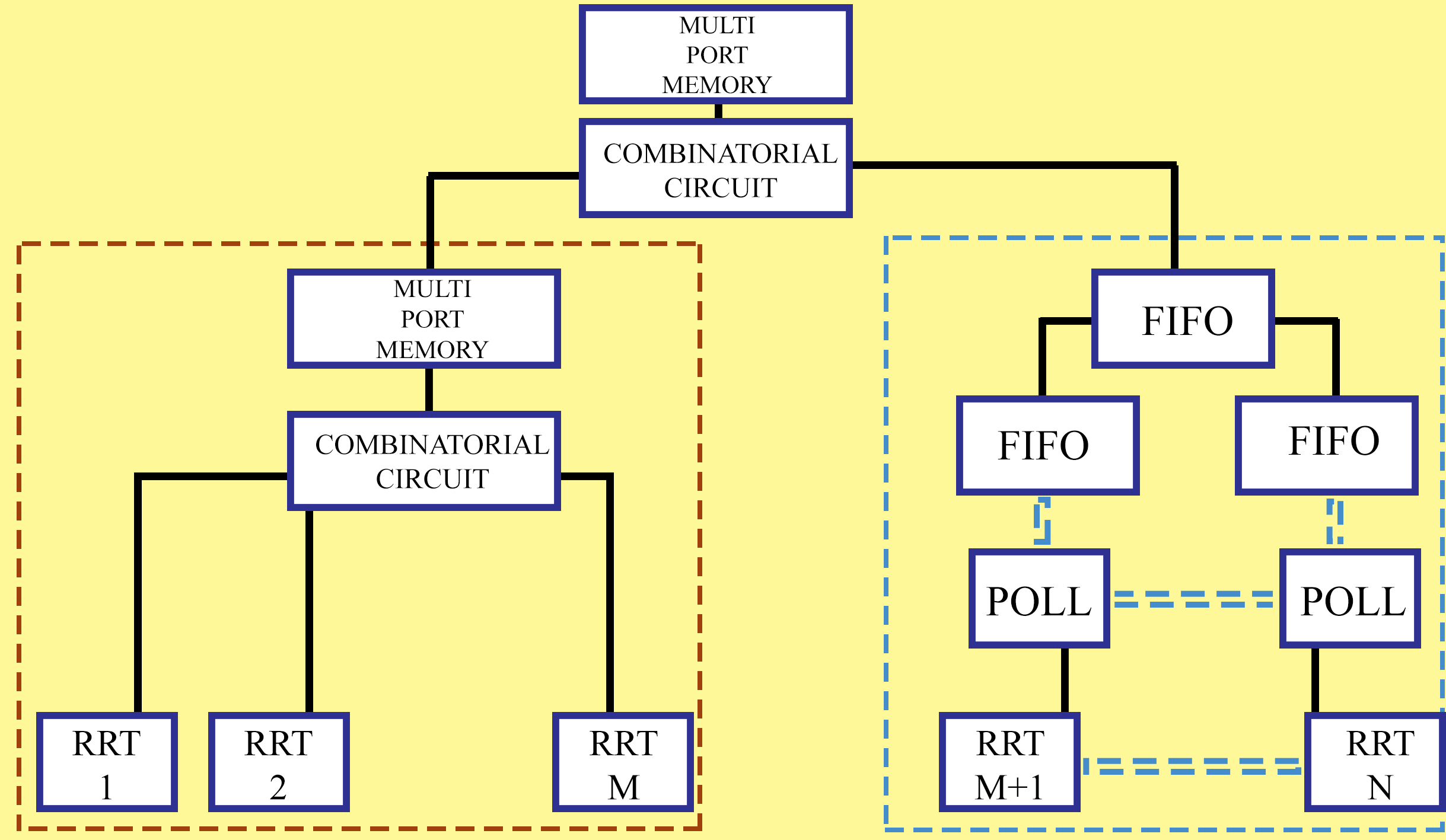}
    \caption{FPGA implementation : Hybrid architecture}
    \label{Hybrid_Architecture2}
\end{figure}

Implementation breakdown, in a bottom to top manner, of each module shown in Fig. \ref{Hybrid_Architecture2} follows.

%Remember that this should be a new image with each module highlighted.

\subsection{RRT Module}
\begin{figure}[h]
    \includegraphics[width=8.6cm, height=3.3cm]{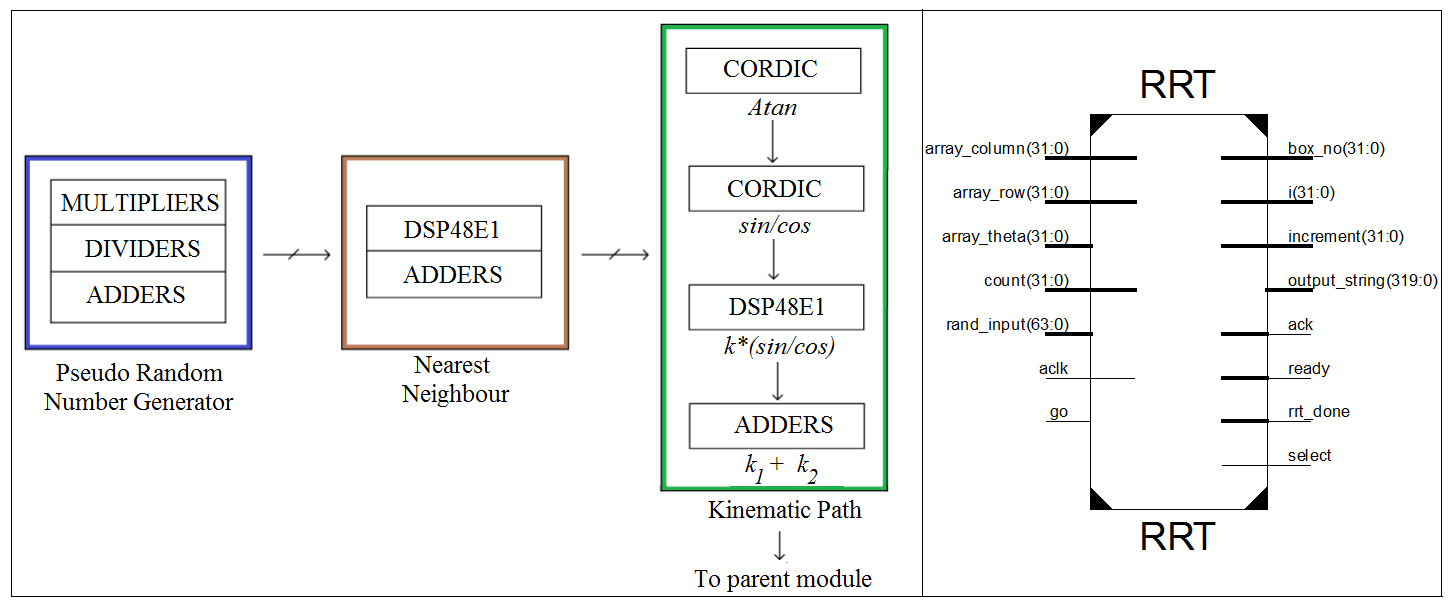}
    \caption{RRT implementation(Best viewed at 500\% zoom)}
    \label{RRT_Schematic}
\end{figure}

As shown in Fig. \ref{RRT_Schematic}, a pseudo-random number generator generates a random state for the mobile robot in use. We use the box \cite{svenstrup2011minimising} method to find the nearest node. Deployment of DSP48E1 slices minimizes the time complexity of distance computation. CORDIC cores are used for computation of trigonometric functions. DSP slices are then used for kinematic extension.

\subsection{POLL}
\begin{figure}[h]
    \includegraphics[width=8.6cm, height=3.3cm]{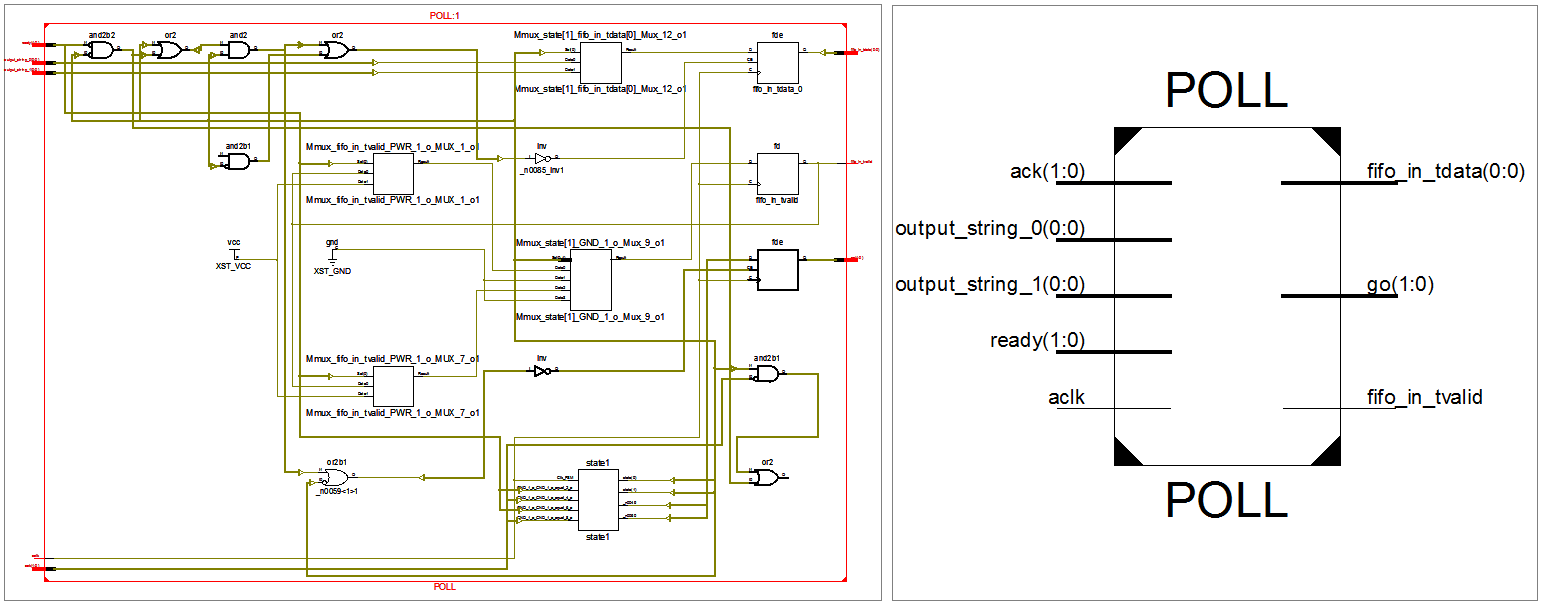}
    \caption{POLL implementation(Best viewed at 500\% zoom)}
    \label{POLL_Schematic}
\end{figure}

 As shown in Fig. \ref{POLL_Schematic}, the POLL is implemented as a sequential Finite State Machine(FSM). Isochronal cyclic polling of child RRT modules germinated by rising edge of clock leads to capture of data bus by one of the children, which then transfers its generated nodes via write-acknowledge mechanism.

\subsection{FIFO}

\begin{figure}[h]
    \includegraphics[width=8.6cm, height=3.0cm]{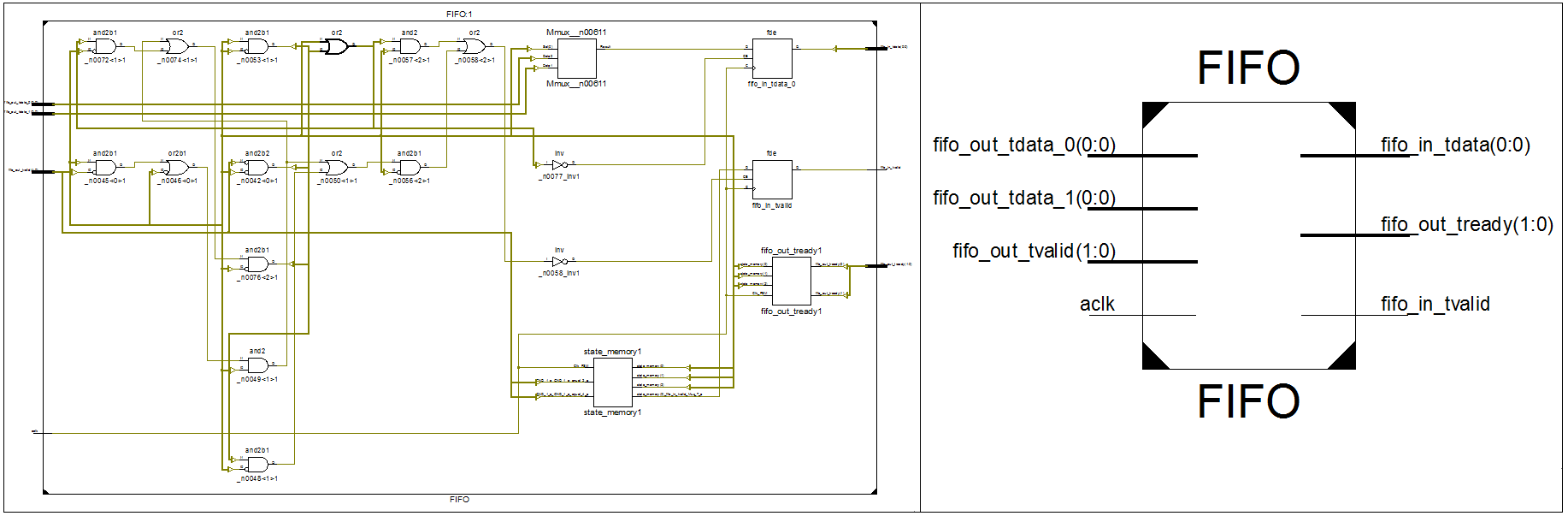}
    \caption{FIFO implementation(Best viewed at 500\% zoom)}
    \label{FIFO_Schematic}
\end{figure}

As Fig. \ref{FIFO_Schematic} shows, we use built in FIFO resources to create high performance, area optimized FIFO module. The First-Word-Fall-Through is chosen as the mode of operation for the FIFO interface.

 \subsection{Combinatorial Circuit}
\begin{figure}[h]
    \includegraphics[width=8.6cm, height=3.1cm]{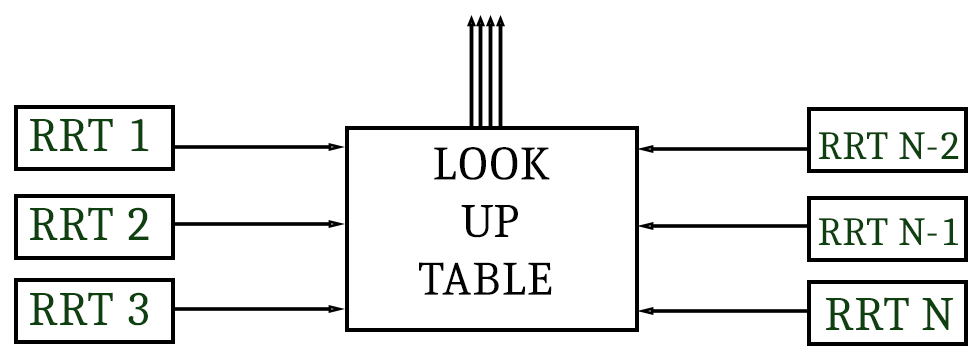}
    \caption{Schematic illustration of combinatorial circuit}
    \label{Combinatorial_Circuit}
\end{figure}

 For N RRTs, the $2^N$ possible cases and the corresponding control signals of the multi-port memory are mapped to cascaded look up tables(LUTs). An N bit string, where each bit corresponds to a RRT, is used as input. A ’1’ bit means that the corresponding RRT is requesting access and a ’0’ bit means otherwise. The outputs of this module are the control signals of the multi-port memory, as shown in Fig. \ref{Combinatorial_Circuit}.

\subsection{Multi-port Memory}
\begin{figure}[h]
    \includegraphics[width=8.6cm, height=3.1cm]{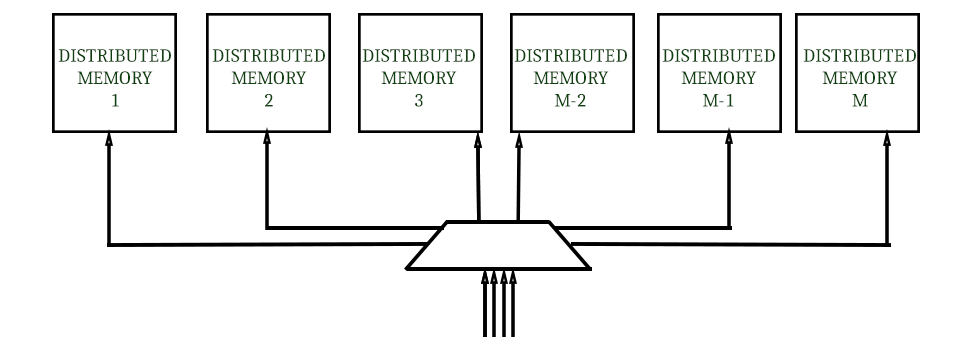}
    \caption{Schematic illustration of multi-port memory}
    \label{Multi-port_Memory}
\end{figure}

With a global address space, the multi-port memory is implemented as a heap of $M$ distributed, single channel memories, each of size $(400∗F)/M$ KB, where $F$ is the number of degrees of freedom of the robot and $M$ is the number of RRTs. The read and write channels are designed asynchronous to enable independent read and write transactions. Auxiliary multiplexers on the read and write channels apportion the global address space to local address spaces, as shown in Fig. \ref{Multi-port_Memory}.

 \section{RESULTS}
\begin{figure}[h]
    \includegraphics[width=8.6cm, height=10.2cm]{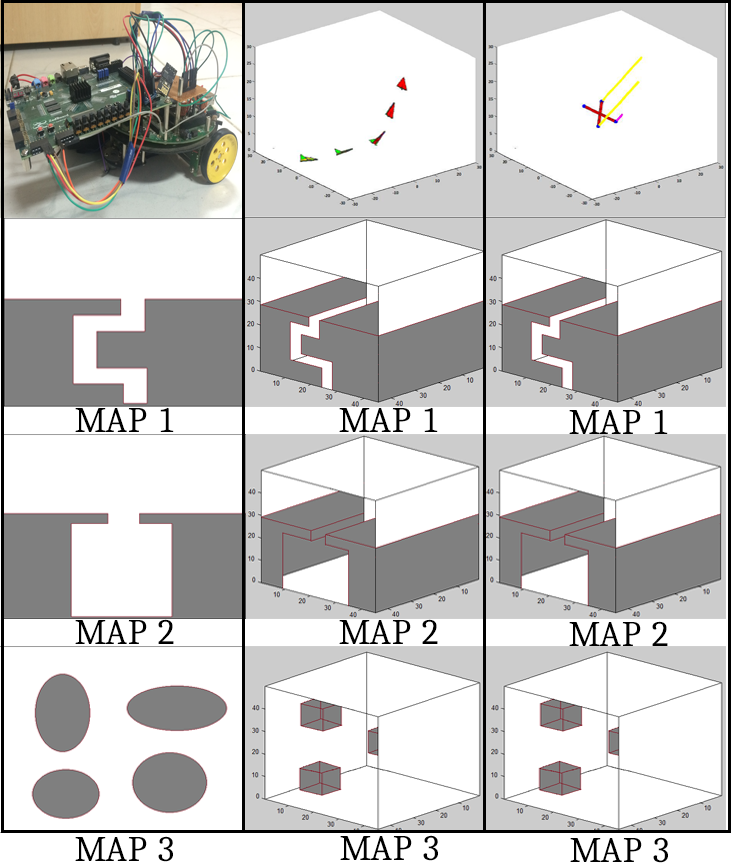}
    \caption{Left Column : Differential, Middle : FWA, Right : Quad-Copter}
    \label{Details_Test_System}
\end{figure}

The experimental setup was planned to qualitatively quantify the architecture across 3 parameters : 1.) Efficiency/performance-per-watt, 2.) Scalability across map's geometric complexity and 3.) Scalability across kinematic complexity. Deployment across 1.) Differentially Steered Firebird V(Actual Run), 2.) Quad-Copter (Simulation) and 3.) Fixed Wing Aircraft (Simulation). Fig. \ref{Details_Test_System} catalogues the test kinematics(first row) and the corresponding 3 maps the architecture was tested on to quantify the above mentioned parameters.

\begin{figure*}[ht]
\centering
\subfigure[Map 1 : Efficiency VS No. of RRTs]{%
    \includegraphics[width=5.4cm, height=3.4cm]{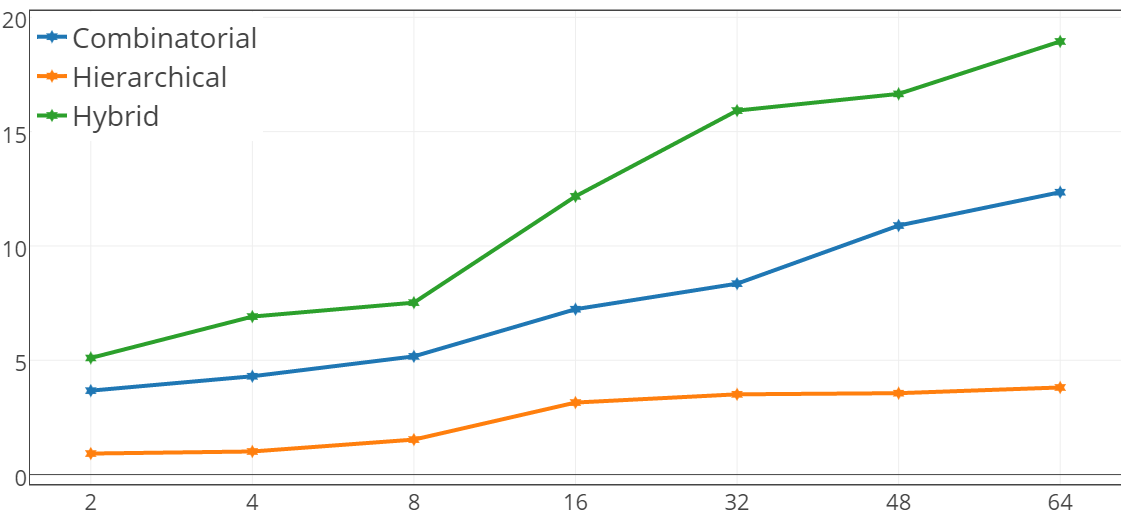}
    \label{Differential_Map1}}
\quad
\subfigure[Map 2 : Efficiency VS No. of RRTs]{%
    \includegraphics[width=5.4cm, height=3.4cm]{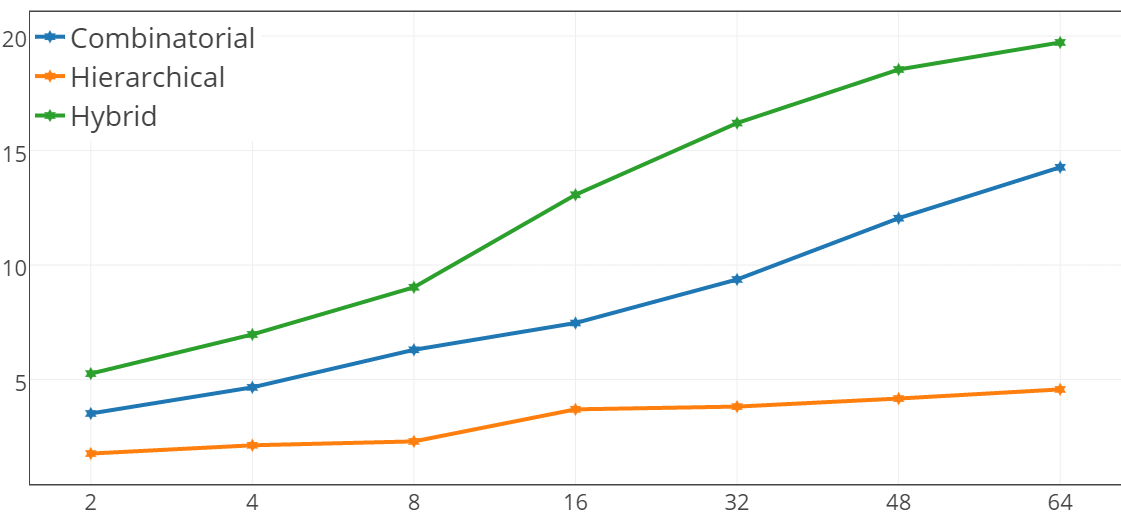}
    \label{Quad_Map1}}    
\quad
\subfigure[Map 3 : Efficiency VS No. of RRTs]{%
    \includegraphics[width=5.4cm, height=3.4cm]{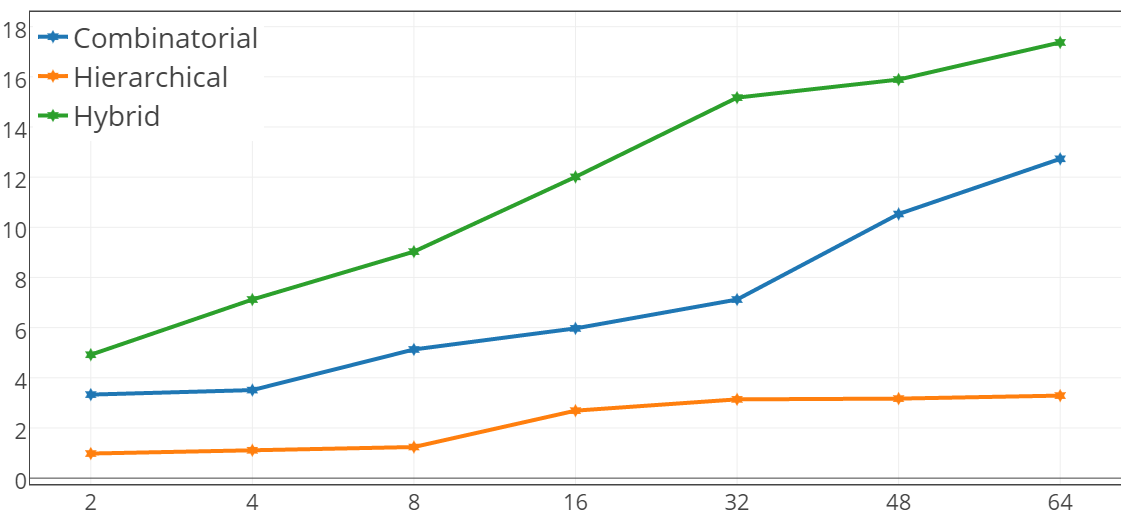}
    \label{FWA_Map1}}    

\subfigure[Map 1 : Efficiency VS No. of RRTs]{%
    \includegraphics[width=5.4cm, height=3.4cm]{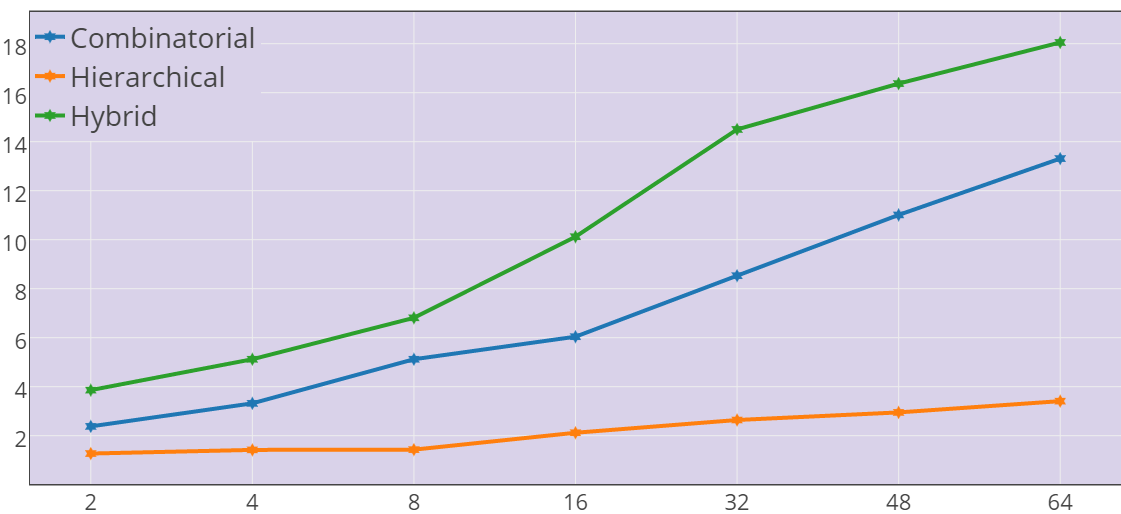}
    \label{Differential_Map2}}
\quad
\subfigure[Map 2 : Efficiency VS No. of RRTs]{%
    \includegraphics[width=5.4cm, height=3.4cm]{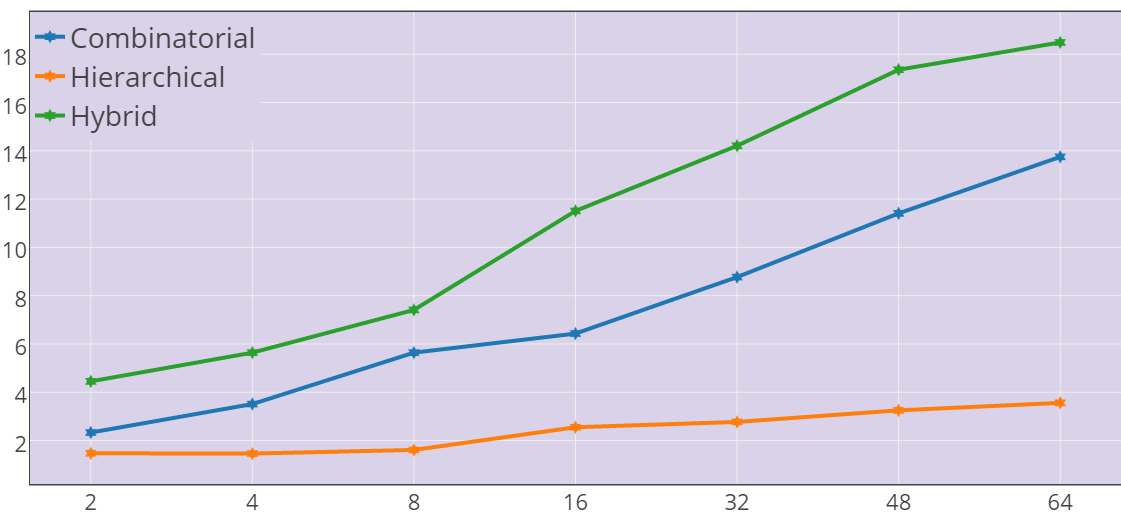}
    \label{Quad_Map2}}    
\quad
\subfigure[Map 3 : Efficiency VS No. of RRTs]{%
    \includegraphics[width=5.4cm, height=3.4cm]{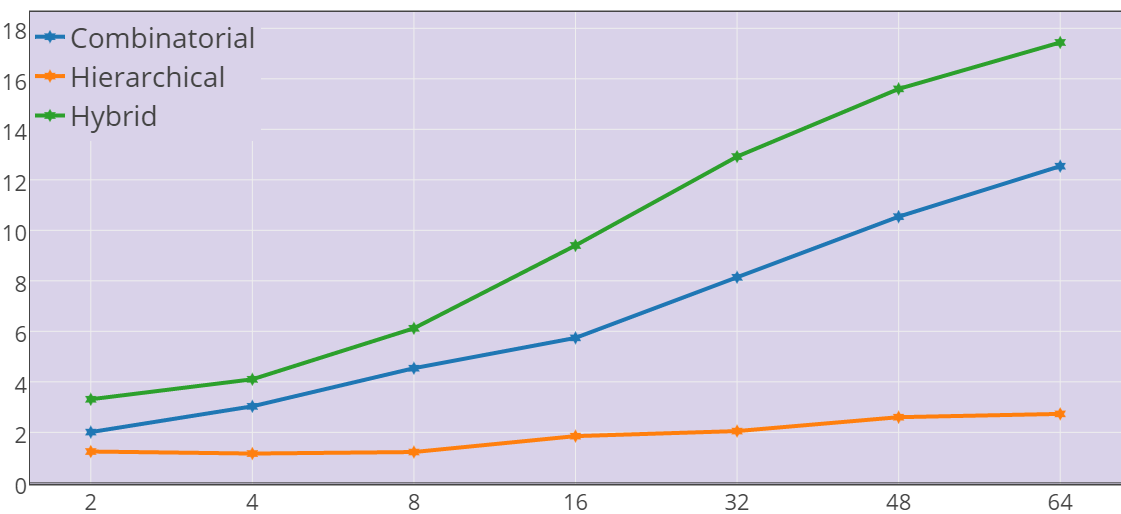}
    \label{FWA_Map2}}    

\subfigure[Map 1 : Efficiency VS No. of RRTs]{%
    \includegraphics[width=5.4cm, height=3.4cm]{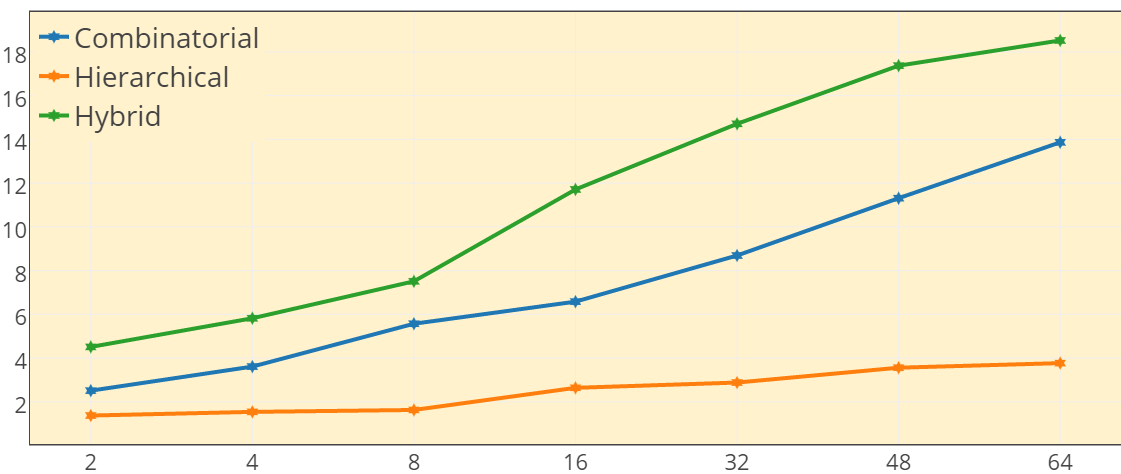}
    \label{Differential_Map3}}
\quad
\subfigure[Map 2 : Efficiency VS No. of RRTs]{%
    \includegraphics[width=5.4cm, height=3.4cm]{FWA_Map2}
    \label{Quad_Map3}}    
\quad
\subfigure[Map 3 : Efficiency VS No. of RRTs]{%
    \includegraphics[width=5.4cm, height=3.4cm]{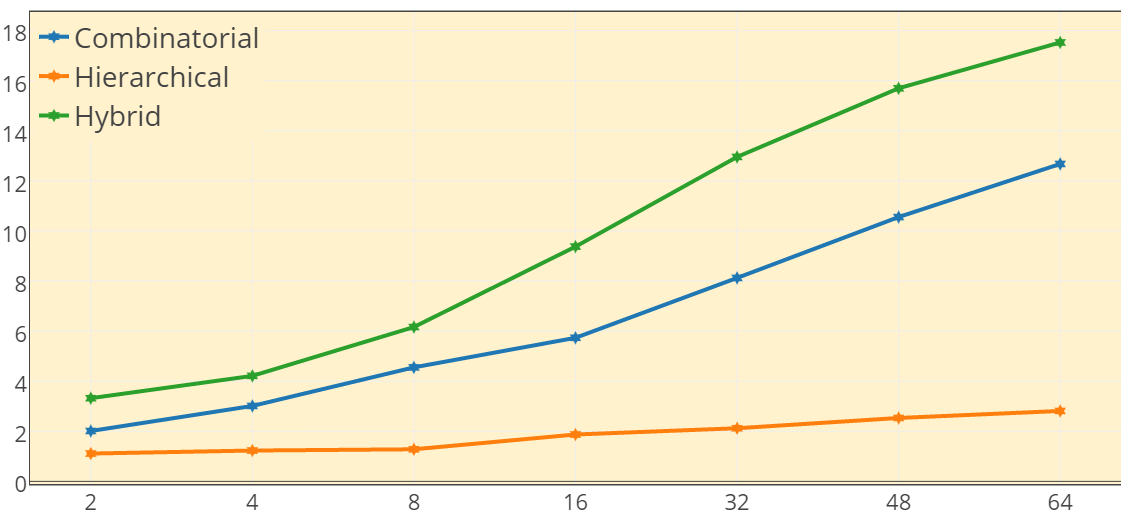}
    \label{FWA_Map3}}

\caption{Comparative analysis of Combinatorial, Hierarchical and Hybrid for the three different test environments across 3 different kinematic test platforms}
\label{Results}
\end{figure*}

The test results quantify 3 parameters across kinematic and geometric complexity : 1.) Speed-Up, 2.) Power consumption and 3.) Performance-per-watt. As described in Eq. \ref{Speed-Up_Equation}, speed-up refers to ratio of the time taken by 1 module to complete a particular task, compared to the time taken by $N$ parallel modules, to complete the same task. Relative to the experimental setup, the equivalent task is to add $10,000$ explored nodes by $N$ parallel RRT modules, initially seeded by a modified K-Means \cite{KMadhava}, to the map. The time is measured by an interrupt driven counter. The power statistics are extracted out of vector-less Vivado power analysis tool. Efficiency or performance-per-watt, as defined in Eq. \ref{Efficiency}, is expected to be the maximum for hybrid architecture.

\begin{equation}
%\scriptsize
S=T(1)/T(N)
\label{Speed-Up_Equation}
\end{equation}
 
\begin{equation}
%\scriptsize
E=S/P
\label{Efficiency}
\end{equation}

\begin{table}[h!]
\centering
\scalebox{0.72}{
\begin{tabular}{|c|c|c|c|c|} 

\hline
\rowcolor{gray!20} MAPS & N=4 & N=16 & N=32 & N=64 \\ [0.5ex] 
\hline
Map 1 & 27, 20.8, 20.4 & 97.6, 82.0, 82.3 & 189.6, 191.3, 190.5 & 424.1, 441.3, 440.1 \\ 
\hline
Map 2 & 28.9, 21.8, 22.3 & 99.4, 85.9, 87.6 & 213.7, 197.0, 198.2 & 472.5, 455.6, 459.1 \\
\hline
Map 3 & 21.5, 18.8, 18.5 & 79.0, 77.5, 77.2 & 162.3, 182.1, 182.3 & 423.2, 420.4, 421.3 \\
\hline
POWER(W) & 6.2, 6.2, 6.3 & 13.4, 13.5, 13.5 & 22.7, 22.4, 22.4 & 34.3, 33.1, 33.3 \\
\hline
\rowcolor{Cyan!20}Map 1 & 2.2, 3.4, 3.5 & 9.7, 6.8, 6.9 & 12.1, 8.5, 8.4 & 17.0, 14.6, 14.5 \\ 
\hline
\rowcolor{Cyan!20} Map 2 & 5.2, 3.7, 3.8 & 11.9, 8.2, 8.5 & 13.4, 9.8, 10.1 & 21, 16.6, 17.3 \\
\hline
\rowcolor{Cyan!20}Map 3 & 2.3, 3.1, 3.2 & 8.6, 6.1, 6.0 & 11.3, 6.8, 6.7 & 14.2, 11.8, 12.0 \\
\hline
\rowcolor{Cyan!20}POWER(W) & 2.1, 2.4, 2.4 & 3.0, 3.2, 3.3 & 3.4, 3.2, 3.1 & 4.4, 4.2, 4.3 \\
\hline
\rowcolor{yellow!20}Map 1 & 20.0, 15.6, 15.6 & 65.2, 58.5, 58.4 & 142.0, 122.7, 122.3 & 323.6, 315.7, 315.6 \\ 
\hline
\rowcolor{yellow!20} Map 2 & 21.3, 17.5, 17.8 & 70.1, 62.0, 62.9 & 144.6, 132.0, 131.8 & 342.6, 322.8, 321.8 \\
\hline
\rowcolor{yellow!20}Map 3 & 19.5, 12.7, 12.8 & 61.7, 54.2, 54.0 & 127.5, 111.0, 110.4 & 317.4, 302.1, 302.4 \\
\hline
\rowcolor{yellow!20}POWER(W) & 2.8, 3.0, 3.0 & 5.3, 5.7, 5.8 & 8.9, 8.4, 8.5 & 17.0, 17.4, 17.3 \\
\hline
\end{tabular}}
\caption{WHITE:Combinatorial,CYAN:Hierarchical,YELLOW:Hybrid}
 \label{Table}
\end{table} 

%\vspace*{-5em}

It should be noted that, owing to the probabilistic nature of RRT, each iteration was performed $1000$ times to get mean values, which are presented in Fig \ref{Results}. Row 1 of Fig. \ref{Results} benchmarks the architecture for differential drive, row 2 for quad-copter and row 3 for fixed wing aircraft, across a diverse spectrum of geometrically complex maps. It should also be remembered that the line plot is for efficiency while the speed-up and power consumption levels are highlighted for each architecture in the plot in form of (Speed-Up, Power). Please note that Table. \ref{Table} details the speed-up and power consumption levels in the following colors : White=Combinatorial, Cyan=Hierarchical and Yellow=Hybrid. The speed-ups are mentioned in the form (Differential(D), Quad-Copter(Q), Fixed Wing(F)). Quantitative numbers and qualitative reasoning behind the same follows.

\subsection{Speed-Up}
As can be concluded from Table. \ref{Table}, the combinatorial architecture, unconstrained from any scheduling between RRT modules, delivers the highest speed-ups across the spectrum of kinematic and geometric complexity of (D=424.1,Q=441.3,F=440.1), (D=472.5,Q=455.6,F=459.1) and (D=423.2,Q=420.4,F=421.3) for $N=64$ for Map 1, 2 and 3 respectively. The hierarchical architecture, coerced by scheduling between RRT modules, comes in at a distant third with the minimum offered speed-ups. The hybrid architecture, designed as an intelligent hybrid of combinatorial and hierarchical so as to achieve speed-ups that are closer to combinatorial, delivers second highest speed-ups of (D=323.6,Q=315.7,F=315.6), (D=342.6,Q=322.8,F=321.8) and (D=317.4,Q=302.1,F=302.4) for $N=64$ for Map 1, 2 and 3 respectively. Qualitatively, this is enabled by the maximization of the cost function that aims to maximise speed-up and minimise power consumption. While the speed-up offered is definitely lower than combinatorial, it can be confidently concluded that the hybrid architecture delivers on the hypothesis of near combinatorial emulation. 

\subsection{Power Consumption}
It should be remembered that the static, vector-less power analysis is independent of kinematic and geometric complexity. Analysis of power consumption data, as given in Table. \ref{Table} in fourth row of each of the colour segments, reveals that hierarchical, owing to the relatively pliant architecture, consumes the least power levels of (D=4.4W,Q=4.2W,F=4.3W) for $N=64$. Combinatorial, owing to its expansive combinatorial blocks, is the most power hungry among the three. Hybrid, on the other hand, tries to closely border hierarchical, expending (D=17.0W,Q=17.4W,F=17.3W) for $N=64$. Qualitative justification behind this behaviour is explicated by the maximization of the cost function that aims to minimise power consumption.   

\subsection{Efficiency}
To appreciate the benchmarking primacy hybrid architecture enables over other architectures, it is important to understand that the architecture was designed to maximise speed-up and minimise power consumption concurrently, despite them being antithetical to each other by nature. The maximization of the designed cost function should enable the hybrid architecture to be the most judicious in efficiency or performance-per-watt. This hypothesis is proven true in Fig. \ref{Results}. As quantized in previous subsections, combinatorial architecture delivers the maximum performance and hierarchical consumes the least amount of power. But, the hybrid tends to closely track the leader in both departments, as already seen. This loose behavioral emulation by the hybrid architecture allows the hybrid architecture to out-throttle both combinatorial and hierarchical in terms of performance-per-watt/efficiency. This out-throttling is observed across the varied spectrum of both geometric as well as kinematic complexities. Hybrid architecture is the most efficient of the 3 architectures with numbers of (D=18.9,Q=18.0,F=18.5), (D=19.7,Q=18.5,F=18.5) and (D=20.4,Q=17,4,F=17.5) for $N=64$ for Map 1, 2 and 3 respectively. This is true across the complete range of $N$.

\begin{figure}[h]
    \includegraphics[width=8.6cm, height=3.4cm]{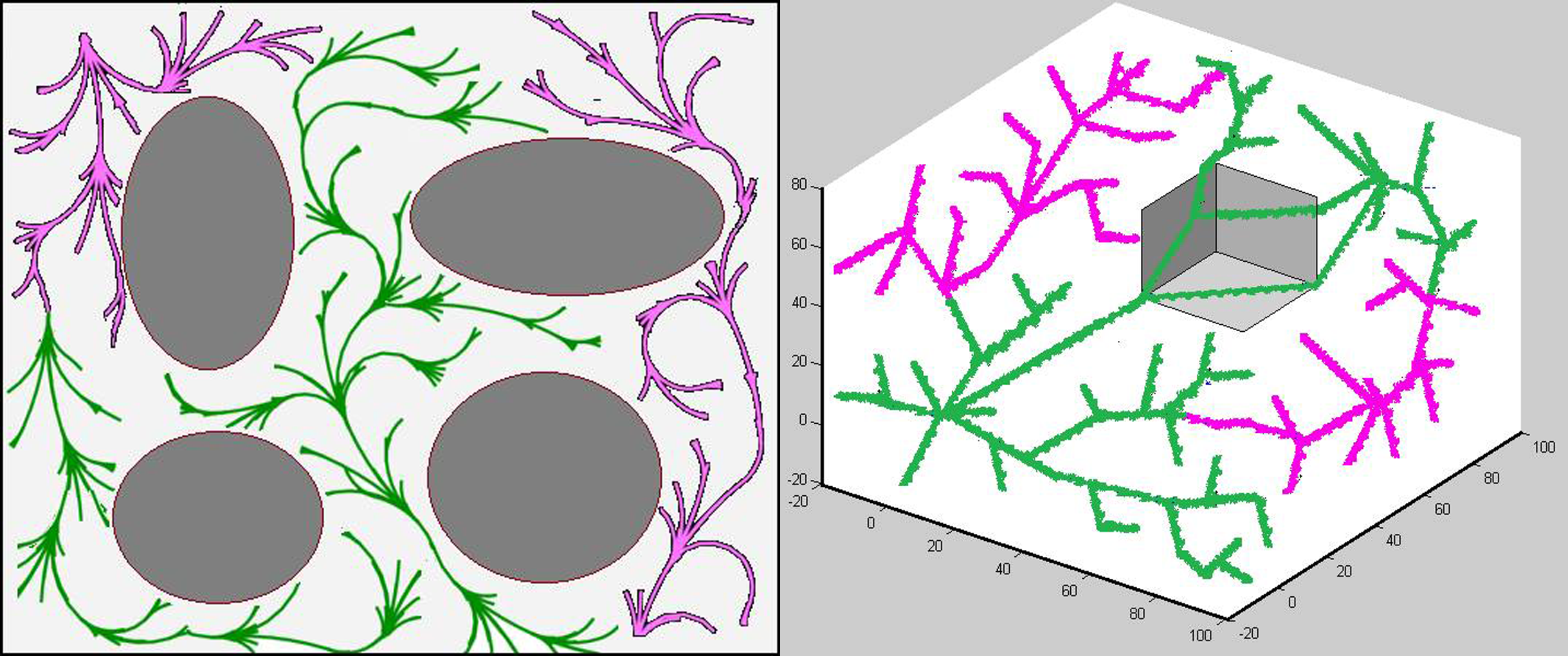}
    \caption{Demo of hybrid architecture for differential and quad-copter}
    \label{Sample_Run}
\end{figure}
 
For perceptible understanding, sample runs of differential drive and quad-copter is presented in Fig. \ref{Sample_Run}. Pink corresponds to exploration by hierarchical and green by combinatorial respectively. Video of demo is also available. This benchmarking exercise, performed across varied kinematics and maps, quantitatively proves the 3 qualities of hybrid architecture : 1.) Maximum efficiency/performance-per-watt, 2.) Scalable across map's geometric complexity and 3.) Kinematic complexity.

\section{CONCLUSION}
This paper proffered the hybrid architecture that, apart from benefiting from the inherent parallel abilities of the FPGA, is able to deliver the maximum performance-per-watt amongst state of the art hardware architectures. Quantitative benchmarking of this architecture across different kinematic systems, from land based kinematics to complex aerial kinematics, on maps with tight geometric constraints exhibited the architecture's scalability across kinematic and geometric complexity.

As part of our future work, the authors would like to study the scalability of this architecture for non-still, dynamically changing maps with moving obstacles. The authors would also like to extend the optimization methods that enables greater combinatorial speed-up and hierarchical power emulation respectively.   

\bibliographystyle{unsrt}
\bibliography{reference}

\end{document}